\documentclass[10pt,twocolumn,letterpaper]{article}

\usepackage{cvpr}              
\definecolor{cvprblue}{rgb}{0.21,0.49,0.74}
\usepackage[pagebackref,breaklinks,colorlinks,allcolors=cvprblue]{hyperref}
\usepackage{multirow}
\usepackage{algorithm}
\usepackage{algorithmic}
\usepackage[table]{xcolor}
\makeatletter
\renewcommand\paragraph{\@startsection{paragraph}{4}{\z@}%
  {2pt}
  {0pt}
  {\normalfont\normalsize\bfseries}}
\makeatother
\def\paperID{37039} 
\def\confName{CVPR}
\def\confYear{2026}

\title{Switch-KD: Visual-Switch Knowledge Distillation for Vision-Language Models}

\author{
Haoyi Sun\thanks{Corresponding author: sunhaoyi@lixiang.com.}, 
Xiaoxiao Wang, 
Ning Mao, 
Qian Wang,\\ 
Lifu Mu,
Wen Zheng,
Tao Wei, 
Wei Chen\\
Li Auto Inc.
}

\begin{document}
\maketitle
\begin{abstract}

Vision-Language Models (VLMs) have shown remarkable capabilities in joint vision–language understanding, but their large scale poses significant challenges for deployment in resource-constrained scenarios.
Knowledge Distillation (KD) offers a viable way to improve model capabilities without increasing model size or data requirements, making deployment more efficient. 
However, applying KD to VLMs is challenged by modality-specific supervision:
although multimodal knowledge in VLMs is fused within the language space, current methods supervise each modality separately without explicitly addressing multimodal alignment, leading to inconsistent multimodal knowledge transfer. 
To address this, we propose \textbf{Switch-KD}, a visual-switch distillation framework that unifies vision–language knowledge transfer within a shared text-probability space. Switch-KD comprises two key components: (1) \textbf{Visual-Switch Distillation}, which switches the student’s visual outputs into the teacher’s language pathway to construct cross-modal probabilistic references for implicit visual knowledge transfer;
and (2) \textbf{Dynamic Bi-directional Logits Difference (DBiLD) loss}, 
which adaptively aligns informative probability regions while preserving the distributional structures of teacher and student through bidirectional supervision. 
Guided by Switch-KD, a 0.5B TinyLLaVA effectively distills rich multimodal knowledge from its 3B teacher, yielding an average improvement of 3.6 points across 10 multimodal benchmarks without any architectural modification. 
Our \href{https://github.com/haoyi199815/Switch-KD}{code} and \href{https://haoyi199815.github.io/Switch-KD}{project page} are publicly available.
\end{abstract}    
\section{Introduction}
\label{sec:intro}

\begin{figure}[htbp]
    \centering
    \includegraphics[width=\linewidth]{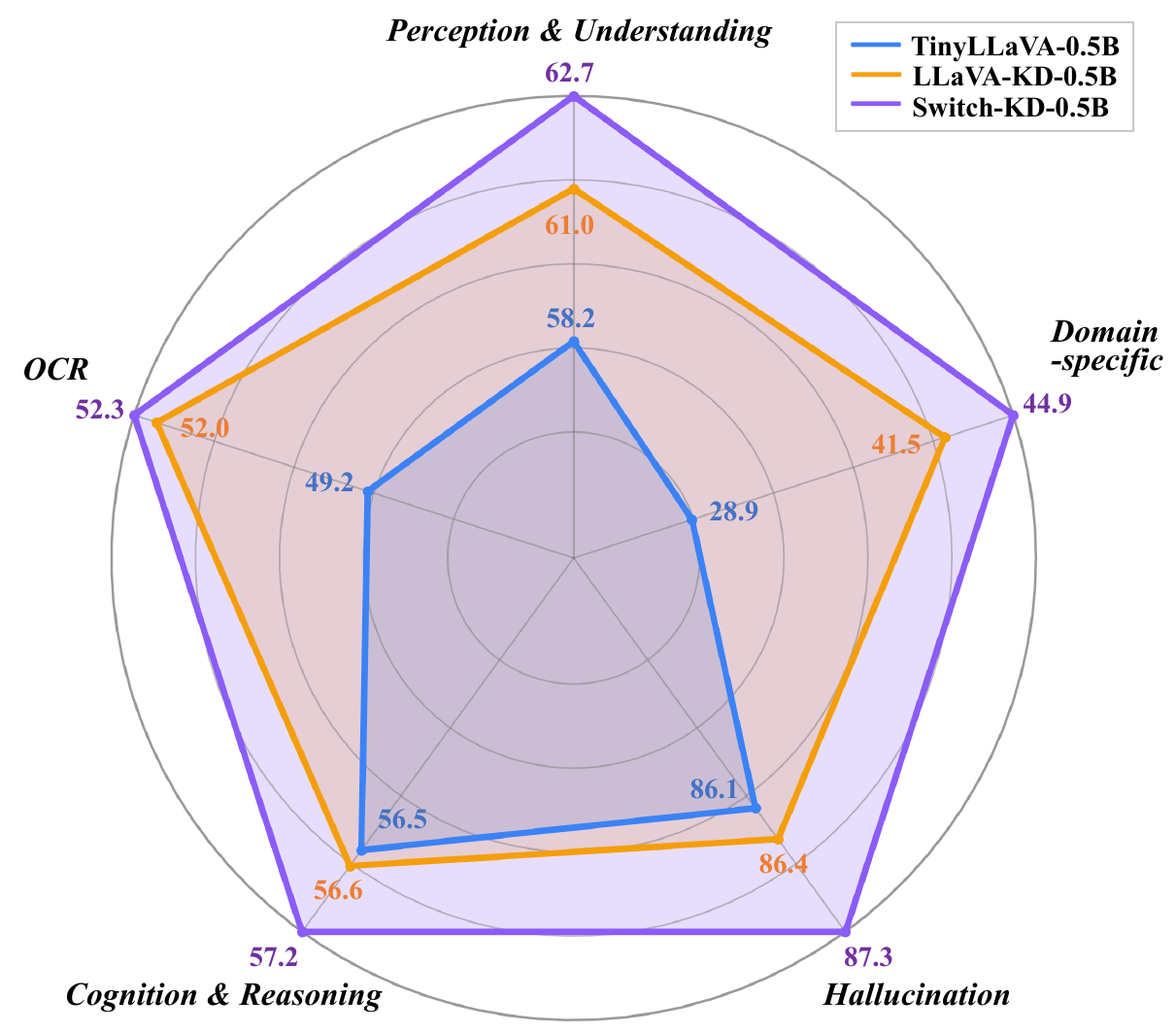}
    \caption{Radar chart comparing the performance of Switch-KD-0.5B, LLaVA-KD-0.5B \cite{cai2024llava}, and TinyLLaVA-0.5B \cite{zhou2024tinyllava} over five competency dimensions. Switch-KD consistently outperforms the other two models, demonstrating its comprehensive and superior capabilities.}
    \label{fig:radar_chart}
\end{figure}


Vision-language models (VLMs) \cite{bai2023qwen,liu2023visual,wang2025internvl3_5} exhibit exceptional performance in joint vision-language understanding, driven by large-scale data and substantial computational resources. Their observed performance improvements closely follow the scaling law. However, the substantial latency, memory footprint, and energy cost of large VLMs limit their applicability in resource-constrained or real-time scenarios, motivating the need for lightweight alternatives.

Knowledge Distillation (KD) \cite{hinton2015distilling,cho2019efficacy,zhao2022decoupled,jin2023multi,huo2024c2kd} offers a well-established paradigm for compressing model scale by transferring knowledge from a high-capacity teacher to a lightweight student, enabling efficient deployment with minimal accuracy degradation. While KD has achieved notable success in text-only language models \cite{gu2023minillm,ko2024distillm,zhang2024dual,li2024bild,wu2025rethinking}, extending it to VLMs remains challenging.

As multimodal knowledge in VLMs is inherently fused within the language space, effective knowledge distillation should likewise transfer cross-modal information through the shared text-probability space. A widely adopted and effective approach, logits distillation \cite{sun2024logit,kim2023token,hao2023one,jin2023multi, zhao2022decoupled}, minimizes the divergence between teacher and student text-probability distributions, serving as strong language-side supervision.
As a natural extension, recent methods have begun to explore supervision on the visual modality alongside the language side. For instance, Align-KD \cite{feng2025align} explicitly aligns respective LLM attentions and respective visual tokens, while LLaVA-KD \cite{cai2024llava} enforces consistency between the LLM-generated visual tokens and their corresponding logits. 
However, these approaches lack tight integration between the visual and language components, employing non-unified supervision across modalities. This non-unified supervision overlooks the intrinsic coupling between modalities, thereby limiting the efficiency of multimodal knowledge transfer.

To address these challenges, we introduce \textbf{Switch-KD}, a unified distillation framework that integrates visual and language knowledge transfer within a shared text-probability space. Switch-KD adopts a novel \textbf{Vision-Switch Distillation} design, switching the student’s visual outputs into the teacher’s language pathway and supervising them with the \textbf{Dynamic Bi-directional Logits Difference (DBiLD) loss}, which adaptively aligns teacher–student distributions via bidirectional Top-$k$ rank matching to enable implicit visual knowledge transfer.

As a key component of Switch-KD, Visual-Switch Distillation integrates two complementary pathways, a standard alignment pathway and a visual-switch pathway, both of which align student–teacher output logits entirely within the shared text-probability space.
The standard pathway follows the conventional approach, matching the student’s logits with those of the teacher.
In contrast, the visual-switch pathway switches the student’s visual encoder outputs into the teacher’s language decoder (S-ViT → T-Projector → T-LLM) to produce visual-switch logits, which represent the teacher’s output distribution conditioned on the student’s encoded visual representations.
This alignment within the shared text-probability space implicitly enables the student to learn from the teacher at the visual-encoder level, achieving visual knowledge transfer through the language pathway.

Effective alignment in the shared text-probability space must address the long-tailed logits of LLMs, as they hinder efficient student imitation during distillation.
BiLD (Bi-directional Logits Difference)~\cite{li2024bild} addresses this issue by bidirectionally aligning the pairwise differences between the teacher’s and student’s top-$k$ logits.
However, a fixed $k$ is incapable of adapting to the different logits distributions observed across different models and samples.
To overcome this limitation, we propose a Dynamic Bi-directional Logits Difference (DBiLD) loss, designed to dynamically select the most informative logits for distillation.
Specifically, we locate the transition point between the information-rich and long-tail regions and adopt it as the adaptive top-$k$ boundary.
Based on this adaptive threshold, DBiLD performs bidirectional top-$k$ alignment between teacher and student logits differences via reverse KL divergence, focusing on high-confidence regions to achieve more stable and effective knowledge transfer.


Our method surpasses both state-of-the-art VLM distillation methods and existing lightweight VLM approaches across a wide range of multimodal benchmarks, as shown in Fig.~\ref{fig:radar_chart}. We summarize our main contributions as follows:

\begin{itemize}
    \item \textbf{Unified text-probability distillation framework.} 
    We propose Switch-KD, a novel distillation framework that unifies multimodal supervision within a shared text-probability space, ensuring consistent knowledge transfer between teacher and student models.

    \item \textbf{Visual-switch distillation with DBiLD loss.} 
    We develop a novel visual-switch distillation architecture that implicitly transfers visual knowledge through the visual-switch pathway, and propose the Dynamic Bi-directional Logits Difference (DBiLD) loss, which adaptively emphasizes informative logits via an adaptive top-$k$ boundary and performs bidirectional alignment with reverse KL divergence.
    
    \item \textbf{Comprehensive empirical validation.} 
    Extensive experiments on 10 multimodal benchmarks show that Switch-KD consistently surpasses existing distillation methods and significantly improves the student’s visual encoder across diverse tasks.
\end{itemize}

\begin{figure*}[htbp]
    \centering
    \includegraphics[width=\linewidth]{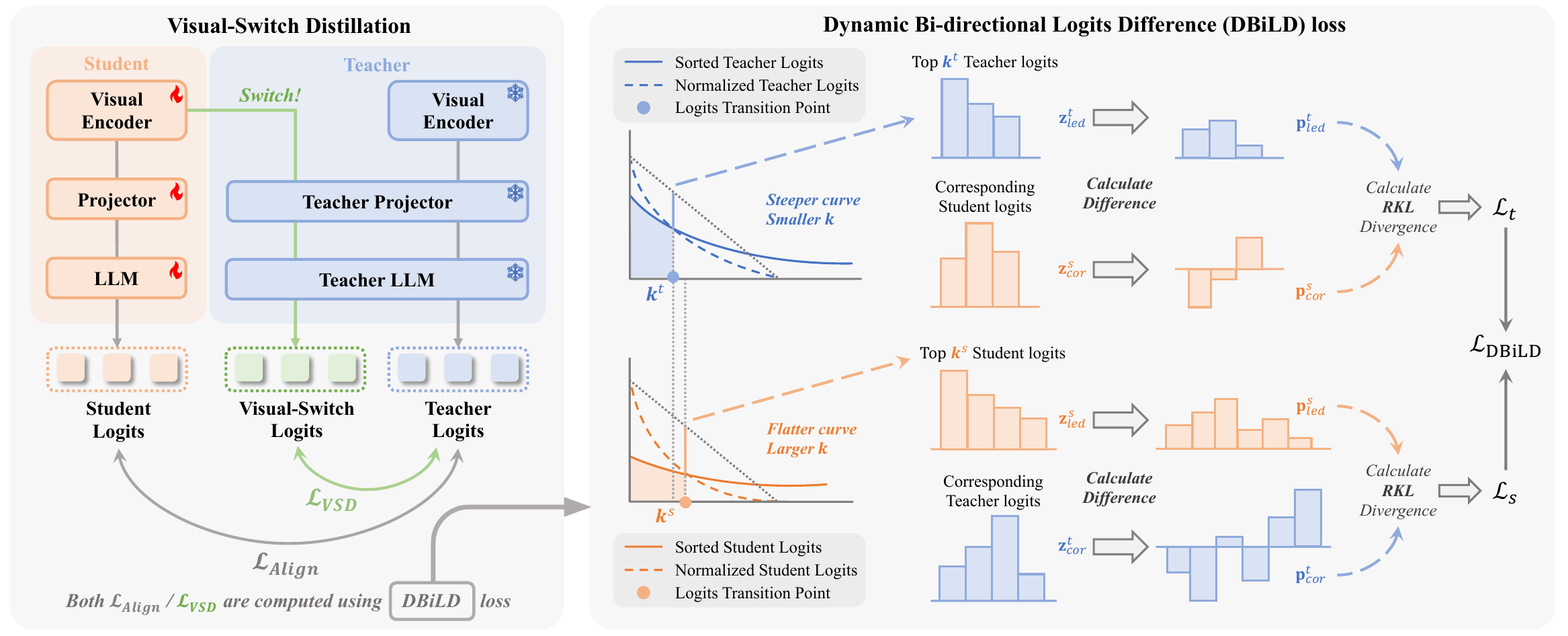}
    \caption{
\textbf{Overview of the proposed Switch-KD framework,}
consisting of two components: 
(a) \textbf{Visual-Switch Distillation} (left), where the student’s visual outputs are switched into the teacher’s language pathway to obtain visual-switch logits for implicit multimodal knowledge transfer; 
and (b) \textbf{DBiLD loss} (right), which first detects knee points $k^t$ and $k^s$ in the respective logits distributions, then constructs two sets of logits differences and bidirectionally aligns them via reverse KL divergence, with an emphasis on high-confidence regions for stable and effective knowledge transfer.
}
    \label{fig:lsda_kd_architecture}
\end{figure*}

\section{Related Work}
\label{sec:formatting}


\paragraph{Vision-Language Models:~}
Recent advances in vision–language models (VLMs) have driven rapid progress in multimodal understanding through increasingly unified architectures \cite{wang2025towards,li2025efficient,li2025bridging,wang2025toward,tian2025watermarking,wang2026star}. Early works such as CLIP \cite{radford2021learning} aligned visual and textual modalities via large-scale contrastive learning, while BLIP-2 \cite{li2023blip} bridged a frozen vision encoder and an LLM using a lightweight Q-Former. Recent models \cite{liu2023visual,bai2023qwen,wang2025internvl3_5} commonly adopt a ViT–Projector–LLM design with instruction tuning for multimodal understanding. 
Scaling analyses~\cite{kumar2024scaling,nezhurina2025scaling,peng2025scaling} demonstrate that VLM performance scales consistently with increases in model size, data volume, and computational budget.
However, in resource-constrained or real-time scenarios, large VLMs suffer from computation and latency bottlenecks, making lightweight alternatives increasingly essential. Recent efforts \cite{zhou2024tinyllava,li2024mini,liu2024sphinx,yao2024minicpm} thus explore compact and efficient architectures that retain competitive performance under limited resources. Building on these advances, Knowledge Distillation (KD) \cite{hinton2015distilling,xu2020feature,wang2021distilling,son2021densely} has emerged as a promising approach to further enhance the performance of lightweight multimodal models.

\paragraph{Knowledge Distillation For VLMs:~} 
KD for large language models (LLMs) have achieved remarkable progress, with most studies \cite{gu2023minillm,ko2024distillm,zhang2024dual,li2024bild,wu2025rethinking} focusing on designing advanced loss functions to effectively transfer textual knowledge from the teacher to the student model.
However, extending KD to VLMs remains challenging due to the additional complexity introduced by the visual modality.
Existing multimodal KD approaches for VLMs can be broadly classified into three categories:
(1) Architecture-enhancement methods. LLaVA-MoD \cite{shu2024llava} and MoVE-KD \cite{cao2025move} extend the student model with Mixture-of-Experts (MoE) structures~\cite{jacobs1991adaptive}, aiming to improve knowledge transfer through sparse parameter activation.
(2) Intermediate-layer supervision methods. VLsI \cite{lee2025vlsi} introduces layer-wise verbalizers to align the student with the teacher’s reasoning trajectory. While effective, such designs substantially increase architectural complexity and training cost.
(3) Explicit visual-constraint methods. 
Align-KD~\cite{feng2025align} aligns the first-layer text–vision attention of the language model, implicitly aligning visual tokens from a structurally matched visual encoder.
LLaVA-KD \cite{cai2024llava} aligns the self-correlation of LLM-generated visual tokens and additionally enforces logit-level consistency for these tokens . 
While explicit visual supervision facilitates visual knowledge transfer, such modality-separated designs limit the overall distillation efficacy.
Going beyond these approaches, our proposed Switch-KD enables collaborative distillation of cross-modal knowledge entirely within a unified text-probability space, ensuring coherent modality-integrated supervision while preserving architectural efficiency.

\section{Method}
Prior work on multimodal knowledge distillation often treats visual features or visual-modality logits as separate supervision targets, leading to fragmented and inconsistent knowledge transfer.
We revisit this problem from a unified probabilistic perspective and ask: if all modalities can be aligned within a shared language probability distribution, can such alignment enable more unified cross-modal knowledge transfer in place of  modality-specific supervision?
Building on this insight, we propose Switch-KD, a unified framework that combines Visual-Switch Distillation for implicit visual knowledge transfer and Dynamic Bi-directional Logits Difference (DBiLD) loss for adaptive probabilistic alignment, as illustrated in Fig.~\ref{fig:lsda_kd_architecture}.

\subsection{Visual-Switch Distillation}
\label{method:vsd}

As shown in the left of Fig.~\ref{fig:lsda_kd_architecture}, the overall process of visual-switch distillation consists of two forward pathways: the standard alignment pathway and the visual-switch pathway.
The overall training objective of the distillation framework is given by:
\begin{equation}
    \mathcal{L} =\mathcal{L}_{\text{CE}} + \lambda_1\mathcal{L}_{\text{Align}} + \lambda_2 \mathcal{L}_{\text{VSD}},
\end{equation}
where $\mathcal{L}_{\text{CE}}$ denotes the auto-regressive language modeling loss,
and $\lambda_1$ and $\lambda_2$ are balancing coefficients that control the relative contributions of the standard alignment distillation term $\mathcal{L}_{\text{Align}}$ and visual-switch distillation term $\mathcal{L}_{\text{VSD}}$.

Both pathways involve a teacher model 
$M^T = (V^T, P^T, L^T)$ and a student model $M^S = (V^S, P^S, L^S)$. Each model follows the same modular structure: a visual encoder $V$, a projector $P$, and a language model $L$. 

\paragraph{Standard Alignment Pathway. }
Given an input image $\mathbf{x}_v$ and a text prompt $\mathbf{x}_t$, the teacher generates its logits $\mathbf{z}^T\in\mathbb{R}^{1\times N}$ through the standard forward process:
\begin{equation}
    \mathbf{z}^T = M^T(\mathbf{x}_v, \mathbf{x}_t) = L^T(P^T(V^T(\mathbf{x}_v)), \mathbf{x}_t),
\end{equation}
where $N$ denotes the vocabulary size of language model $L^T$.
Similarly, the student model generates the corresponding logits $\mathbf{z}^S\in\mathbb{R}^{1\times N}$ as:
\begin{equation}
    \mathbf{z}^S = M^S(\mathbf{x}_v, \mathbf{x}_t) = L^S(P^S(V^S(\mathbf{x}_v)), \mathbf{x}_t).
\end{equation}

During distillation, the teacher is frozen while the student is fully trainable. We minimize the distributional difference between $\mathbf{z}^T$ and $\mathbf{z}^S$ as:
\begin{equation}
    \mathcal{L}_{\text{Align}} = \mathcal{L}_{\text{DBiLD}}(\mathbf{z}^T,\mathbf{z}^S),
\end{equation}
where $\mathcal{L}_{\text{DBiLD}}(\cdot,\cdot)$ represents the general distribution alignment metric detailed in Section \ref{subsec:loss}. This pathway conducts overall supervision in the language space, with comparatively less supervision for the visual modality, motivating the introduction of the visual-switch pathway.

\paragraph{Visual-Switch Pathway. }
To complement the overall text-probability space supervision, the visual-switch pathway explicitly leverages the switch mechanism of $V^S$ for more direct and effective visual knowledge transfer.

The core hypothesis is that if the student visual encoder $V^S$ learns meaningful representations, 
its visual representations $\mathbf{h}_v^S=V^S(\mathbf{x}_v)$ should be correctly interpreted and decoded by the teacher’s language pathway $(P^T, L^T)$, yielding a probability distribution consistent with the teacher’s original output. 
Based on this intuition, we defined the visual-switch logits $\mathbf{z}^{Switch}\in\mathbb{R}^{1\times N}$ as follows:
\begin{equation}
    \mathbf{z}^{Switch} = L^T(P^T(V^S(\mathbf{x}_v)), \mathbf{x}_t),
\end{equation}
where the output of the student’s visual encoder is switched into the frozen teacher projector and language model to form a hybrid inference route, representing the teacher’s output distribution conditioned on the student’s visual representations $\mathbf{h}_v^S$.

To further facilitate implicit visual knowledge transfer in the shared text-probability space, we minimize the visual-switch logits’ discrepancy with the teacher logits:
\begin{equation}
    \mathcal{L}_{\text{VSD}} = \mathcal{L}_{\text{DBiLD}}(\mathbf{z}^T,\mathbf{z}^{Switch}).
\end{equation}

Intuitively, this process can be viewed as a hypothetical scenario: the teacher’s ``brain'' $(P^T, L^T)$ attempts to interpret the world through the student’s ``eyes'' $(V^S)$. Such supervision encourages $V^S$ to produce visual representations that can be readily interpreted by the teacher’s language pathway, thereby enabling implicit visual knowledge transfer within the unified text-probability space.

\subsection{Dynamic Bi-directional Logits Difference Loss}
\label{subsec:loss}

The right part of Fig.~\ref{fig:lsda_kd_architecture} depicts the components of Dynamic Bi-directional Logits Difference (DBiLD) loss, which consists of two complementary branches: a teacher-guided distillation term $\mathcal{L}_t$ and a student-guided term $\mathcal{L}_s$. 
The overall loss is defined as the combination of the two branches:

\begin{equation}
    \mathcal{L}_{\text{DBiLD}} = \mathcal{L}_t + \mathcal{L}_s.
    \label{eq:dbild_main}
\end{equation} 

\paragraph{Teacher-Guided Distillation. }
The teacher-guided component $\mathcal{L}_t$ transfers the teacher’s most confident knowledge to the student by focusing on its most informative logits.
Instead of using a fixed $k$ for all samples, we adopt the Kneedle algorithm~\cite{satopaa2011finding} to adaptively determine $k^t$ for each instance based on the distribution of teacher logits.
The dynamic top-$k$ selection procedure can be described mathematically as follows.

First, given the teacher logits $\mathbf{z}^t\in\mathbb{R}^{1\times N}$ for DBiLD, 
let $\overline{\mathbf{z}}^t$ denote the corresponding logits sorted in descending order. 
We then normalize both their ranks and values to the range $[0, 1]$ to prepare for knee-point detection, yielding:
\begin{equation}
    x_i = \frac{i}{N}, \quad 
    y_i = \frac{\overline{\mathbf{z}}^t_i - \overline{\mathbf{z}}^t_{\min}}{\overline{\mathbf{z}}^t_{\max} - \overline{\mathbf{z}}^t_{\min}}, \quad i = 1, \dots, N.
\end{equation}

Next, we define a reference line $r(x) = 1 - x$ and compute the perpendicular distance $d_i=(1-x_i)-y_i$ between each normalized point $(x_i, y_i)$ and the reference line.
The rank index $i$ with the maximum distance is identified as the dynamic cutoff $k^t$:
\begin{equation}
    k^t = \arg\max_i d_i,
\end{equation}
which represents the transition point where the distribution shifts from the information-rich region to the long-tail region.
After obtaining the dynamic cutoff $k^t$, we selected top-$k^t$ teacher-led logits  $\mathbf{z}^t_{led}$ from $\overline{\mathbf{z}}^t$:
\begin{equation}
    \mathbf{z}^t_{led} = [z^t_{i_1}, z^t_{i_2}, \dots, z^t_{i_{k^t}}]\in\mathbb{R}^{1\times {k^t}}.
\end{equation}
Then, we create the corresponding student logits $\mathbf{z}^s_{cor}$ by selecting the student logit values at the corresponding indices $[i_1,i_2,\dots,i_{k^t}]$:
\begin{equation}
    \mathbf{z}^s_{cor} = [z^s_{i_1}, z^s_{i_2}, \dots, z^s_{i_{k^t}}]\in\mathbb{R}^{1\times {k^t}}.
\end{equation}

Internal pairwise differences are then computed to capture the relative ranking structure:

\begin{equation}
\begin{split}
    \mathbf{d}_{led}^{t} &= [\,z^t_{i_m} - z^t_{i_n} \mid 1 \le m < n \le k^t\,]\in\mathbb{R}^{1\times\frac{k^t(k^t-1)}{2}}, \\
    \mathbf{d}_{cor}^{s} &= [\,z^s_{i_m} - z^s_{i_n} \mid 1 \le m < n \le k^t\,]\in\mathbb{R}^{1\times\frac{k^t(k^t-1)}{2}}.
\end{split}
\end{equation}

The difference vectors are further normalized into probability distributions via temperature-scaled softmax:
\begin{equation}
\begin{split}
    \mathbf{p}_{led}^{t} = \frac{\exp(\mathbf{d}^t_{led} / \tau)}{\sum_{i=1}^\frac{k^t(k^t-1)}{2} \exp(d_{led,i}^t / \tau)}, \\
    \mathbf{p}_{cor}^{s} = \frac{\exp(\mathbf{d}^s_{cor} / \tau)}{\sum_{i=1}^\frac{k^t(k^t-1)}{2} \exp(d_{cor,i}^s / \tau)}.
\end{split}
\end{equation}
where $\tau$ is the temperature parameter controlling distribution sharpness.

The teacher-guided loss is then computed using reverse KL divergence (RKL) to align the student’s distribution with the teacher’s:
\begin{equation}
    \mathcal{L}_t = D_{\text{RKL}}[\mathbf{p}_{led}^{t} \,\|\, \mathbf{p}_{cor}^{s}],
\end{equation}
which distills knowledge from the high-confidence region of the teacher distribution through a targeted focus on the most informative logits determined by $k^t$.

\paragraph{Student-Guided Distillation. }
As a counterpart to the teacher-guided branch, the student-guided distillation component $\mathcal{L}_s$ 
performs the same bidirectional alignment process, but from the student’s perspective. 
Instead of using the teacher logits to identify the high-confidence region, 
we dynamically select the top-$k^s$ logits based on the student distribution, 
where $k^s$ is obtained via the same knee-point detection procedure applied to $\mathbf{z}^s$. 
This adaptive selection enables the student to verify its most confident predictions against the corresponding teacher logits.

Analogously, we extract the top-$k^s$ student-led logits $\mathbf{z}_{led}^{s}$  which sorted in descending order along with the corresponding teacher logits $\mathbf{z}_{cor}^{t}$ at the same indices. Based on these, we can sequentially compute their internal pairwise differences $\mathbf{d}_{led}^{s}$ and $\mathbf{d}_{cor}^{t}$ as well as the probability distributions $\mathbf{p}_{led}^{s}$ and $\mathbf{p}_{cor}^{t}$.


The student-guided loss is then expressed as:
\begin{equation}
    \mathcal{L}_s = D_{\text{RKL}}[\mathbf{p}_{cor}^{t} \,\|\, \mathbf{p}_{led}^{s}],
\end{equation}
which verifies the student’s most confident predictions against the teacher’s distribution, thereby complementing the teacher-guided branch and establishing bidirectional symmetric supervision between the two models.

\section{Experiments}
\label{exp}
\begin{table*}[ht]
\centering
\scriptsize
\setlength{\tabcolsep}{3pt}
\renewcommand{\arraystretch}{1.15}
\begin{tabular}{lllccccccccccccc}
\toprule
\multirow{2}{*}{\textbf{Method}} & 
\multirow{2}{*}{\textbf{LLM}} & 
\multirow{2}{*}{\#\textbf{Samples}} & 
\multirow{2}{*}{\textbf{Distillation}} &
\multicolumn{3}{c}{\textbf{Percep. \& Underst.}} &
\multicolumn{4}{c}{\textbf{Cognition \& Reasoning}} &
\multicolumn{1}{c}{\textbf{OCR}} &
\multicolumn{1}{c}{\textbf{Specific}} &
\multicolumn{1}{c}{\textbf{Halluc.}} &
\multirow{2}{*}{\textbf{Avg\textsubscript{7}}} &
\multirow{2}{*}{\textbf{Avg\textsubscript{10}}}\\
\cmidrule(lr){5-7}\cmidrule(lr){8-11}\cmidrule(lr){12-12}\cmidrule(lr){13-13}\cmidrule(lr){14-14}
& & & & MME & MMB & MMB$^{\text{CN}}$ & VQAv2 & GQA & SciQA & MMMU & TextVQA & VizWiz & POPE & &\\
\midrule
LLaVA-1.5 & Vicuna-7B & 1.2M && 75.5 & 64.3 & 58.3 & 78.5 & 62.0 & 66.8 & 34.4 & 58.2 & 50.0 & 85.9 & 62.2 & 63.4\\
MoVE-KD-v1.1 &Vicuna-7B&1.2M& \checkmark &75.5&67.4&--&79.9&63.9&69.8&--&59.6&52.7&86.3&--&--\\
Qwen-VL-Chat & Qwen-7B & 1500M & & 74.4 & 60.6 & 56.7 & 78.2 & 57.5 & 68.2 & 35.9 & 61.5 & 38.9 & -- & 59.7 & --\\
TinyLLaVA$^\dagger$ & Qwen2.5-7B & 1.2M &  &
77.4 & 74.9 & 74.4 & 81.3 & 64.0 & 73.6 & 41.6 & 60.3 & 53.9 & 86.8 &68.4 &68.8\\
TinyLLaVA$^\dagger$ & Qwen2.5-3B & 1.2M &  &
73.9 & 71.8 & 69.5 & 80.4 & 63.2 & 76.0 & 40.3 & 61.5 & 38.7 & 86.4 & 64.9 & 66.2\\

\midrule
\rowcolor{gray!10}
TinyLLaVA$^\dagger$ & Qwen2.5-1.5B & 1.2M & &
\textbf{72.5} & 68.6 & 63.0 & 78.8 & 62.0 & \textbf{72.0} & \textbf{37.0} & 57.4 & 43.2 & 85.5 & 62.7 & 64.0\\
Mini-Gemini-2B&Gemma-2B&2.7M& &
67.0&59.8&51.3&--&60.7&63.1&31.7&56.2&41.5&85.6&57.1&--\\
MoVE-KD-v1.1 & MobileLLaMA-1.4B & 1.2M & \checkmark &
59.4 & 48.8 & -- &73.8 &57.7&57.3&--&44.3&29.3&86.1&--&--\\
LLaVA-MOD & Qwen1.5-1.8B & 5M & \checkmark&
66.7 & 66.3 & 61.9 & -- & 58.7 & 68.0 & -- & 58.5 & 39.2 & \underline{87.0} & 59.9 & --\\
\rowcolor{blue!10}
LLaVA-KD & Qwen2.5-1.5B & 1.2M & \checkmark &
70.0 & \underline{71.0} & \underline{66.6} & \underline{80.3} & \underline{62.5} & \underline{71.6} & \underline{35.8} & \underline{59.7} & \textbf{46.0} & 86.7 & \underline{63.9} & \underline{65.0}\\
\rowcolor{red!10}
Switch-KD & Qwen2.5-1.5B & 1.2M& \checkmark &
\underline{72.2} & \textbf{71.4} & \textbf{68.5} &\textbf{81.4}&\textbf{63.9} &69.3&34.9&\textbf{60.3}&\underline{44.4}&\textbf{86.8}&\textbf{64.3}&\textbf{65.3}\\
\midrule
\rowcolor{gray!10}
TinyLLaVA$^\dagger$ & Qwen2.5-0.5B & 1.2M & &
61.5 & 58.9 & 54.2 & 74.8 & 58.3 & 59.1 & \textbf{33.6} & 49.2 & 28.9 & 86.1 & 52.9 & 56.5\\
SPHINX-Tiny & TinyLlama-1.1B &15M & &
63.1& 52.3& 56.6&74.7&58.0&21.5&31.1&\textbf{57.8}&\textbf{49.2}&82.2&51.2&54.7\\
LLaVA-MOD & Qwen1.5-0.5B & 5M & \checkmark&
\underline{65.3} & 58.8 & 50.4 & -- & 56.2 & \textbf{62.8} & -- & \underline{53.9} & 31.6 & -- & 54.1 & --\\
\rowcolor{blue!10}
LLaVA-KD & Qwen2.5-0.5B & 1.2M & \checkmark&
64.7 & \underline{61.3} & \underline{57.0} & \underline{77.7} & \underline{59.8} & \underline{60.6} & 28.3 & 52.0 & 41.5 & \underline{86.4} & \underline{56.7} & \underline{58.9}\\
\rowcolor{red!10}
Switch-KD & Qwen2.5-0.5B & 1.2M& \checkmark&
\textbf{66.8}&\textbf{63.5}&\textbf{57.8}&\textbf{79.6}&\textbf{61.6}&57.9&\underline{29.8}&52.3&\underline{44.9}&\textbf{87.3}&\textbf{57.8}&\textbf{60.1}\\
\bottomrule
\end{tabular}
\caption{
\textbf{Benchmark results of Switch-KD vs. state-of-the-art VLMs.}
Compared with counterparts, our Switch-KD consistently achieves superior or comparable results across various benchmarks. 
Optimal and sub-optimal results are highlighted in \textbf{bold} and \underline{underline}.
\textcolor{gray!70}{Grey}, \textcolor{blue!50}{blue}, and \textcolor{red!50}{red} backgrounds denote our baseline models, most relevant KD methods, and our method, respectively.
$\text{Avg}_{7}$ represents the average over seven benchmarks (excluding VQAv2, POPE, and MMMU),
while $\text{Avg}_{10}$ averages over all benchmarks.
$^\dagger$: reproduced results using the official code.
}
\label{tab:Switch-kd}
\end{table*}
\begin{table*}[t]
\centering
\small
\renewcommand{\arraystretch}{1.15}
\setlength{\tabcolsep}{4pt}
\begin{tabular}{l l l c c c c c c c c}
\toprule
\textbf{Subset / \#Samples} & \textbf{Student LLM} & \textbf{Method}   & \textbf{MME$^{P}$} & \textbf{MMB} & \textbf{GQA} & \textbf{SciQA} & \textbf{TextVQA} & \textbf{POPE} & \textbf{Avg.} & \textbf{$\Delta$} \\
\midrule
\multirow{2}{*}{\textit{Short} / 3.6M} 
& MobileLLaMA-1.7B & MobileVLM V2 & 1246.3 & 57.6 & 55.1 & 63.2 & 51.2 & 85.3   & 62.4 & - \\
& MobileLLaMA-1.7B & Align-KD & 1288.4 & 57.8 & 58.9 & 66.6 & 52.4 & 86.5  & 64.4 & 2.0 \\
\midrule
\multirow{2}{*}{\textit{Long} / 3.6M} 
& MobileLLaMA-1.7B & MobileVLM V2 & 1289.2 & 55.9 & 59.0 & 64.5 & 52.2 & 86.1  & 63.7 & - \\
& MobileLLaMA-1.7B & Align-KD & 1303.8 & 57.5  & 60.1 & 67.7 & 53.1 & 87.0  & 65.1 & 1.4 \\
\midrule
\multirow{2}{*}{\textit{Short} / 1.2M}& \cellcolor{gray!10}Qwen2.5-1.5B & \cellcolor{gray!10}MobileVLM V2 
& \cellcolor{gray!10}1365.9 
& \cellcolor{gray!10}65.6 
& \cellcolor{gray!10}58.0 
& \cellcolor{gray!10}68.9
& \cellcolor{gray!10}51.2 
& \cellcolor{gray!10}84.2  
& \cellcolor{gray!10}66.0 
& \cellcolor{gray!10}-\\
& \cellcolor{red!10}Qwen2.5-1.5B 
& \cellcolor{red!10}Switch-KD
& \cellcolor{red!10}\textbf{1411.5} 
& \cellcolor{red!10}\textbf{68.4}
& \cellcolor{red!10}\textbf{61.9} 
& \cellcolor{red!10}\textbf{71.6}
& \cellcolor{red!10}\textbf{57.0} 
& \cellcolor{red!10}\textbf{87.5}  
& \cellcolor{red!10}\textbf{69.5} 
& \cellcolor{red!10}\textbf{3.5}\\

\bottomrule
\end{tabular}
\caption{
\textbf{Performance comparison of Switch-KD with Align-KD.} Using fewer training samples and a lighter LLM backbone, Switch-KD achieves superior performance over the baseline. 
MME$^{P}$ refers to MME Perception.
\textit{Short} and \textit{Long} refer to two subsets with different maximum prompt lengths limitations. $\Delta$ denotes the average improvement of the distillation method over its baseline.
}
\label{tab:align_kd_results}
\end{table*}

\subsection{Setup}
\paragraph{Setting. }
To ensure a fair comparison in Table~\ref{tab:Switch-kd}, our setting largely inherits the training configuration of LLaVA-KD~\cite{cai2024llava}, employing the pretrained SigLIP-B/14@384px \cite{zhai2023sigmoid} as the visual encoder, a two-layer MLP with GELU activation as the projector and Qwen2.5 family models as the LLM backbone. 
For comparison with Align-KD~\cite{feng2025align}, as shown in Table~\ref{tab:align_kd_results}, we also conduct experiments adopting CLIP ViT-L/14~\cite{radford2021learning} as the visual encoder, LDPv2 \cite{chu2024mobilevlm} as the projector, and a lighter Qwen2.5-1.5B backbone instead of the 1.7B LLM~\cite{chu2023mobilevlm} used in the original Align-KD.
We divide the training process into two stages: Pre-Training (PT) and Distilled Fine-Tuning (DFT). The PT stage uses LLaVA1.5-558K~\cite{liu2024improved} to establish vision–language alignment, while DFT leverages LLaVA-Mix-665K~\cite{liu2024improved} for effective knowledge distillation.
For both stages, the teacher and student models share the same overall architecture. 
Specifically, the 0.5B student model is distilled from a teacher with a Qwen2.5-3B LLM backbone, 
while the 1.5B student model is distilled from a teacher with a Qwen2.5-7B LLM backbone. The temperature parameter $\tau$ is setted as 3.
Loss weights for $\mathcal{L}_{\text{Align}}$ and $\mathcal{L}_{\text{VSD}}$ are consistently set to $\lambda_1=1.0$ and $\lambda_2=1.0$ in all experiments.
\paragraph{Benchmarks. }
We evaluate Switch-KD on ten widely used multimodal benchmarks, organized into five categories~\cite{li2024survey}:
(1) Perception \& Understanding: MMB~\cite{liu2024mmbench}, MMB\textsuperscript{CN}~\cite{liu2024mmbench}, and MME~\cite{yin2024survey} evaluate general visual–language understanding, multilingual perception, and multimodal alignment capabilities.
(2) Cognition \& Reasoning: VQAv2~\cite{goyal2017making}, GQA~\cite{hudson2019gqa}, ScienceQA~\cite{lu2022learn}, and MMMU~\cite{yue2024mmmu} evaluate compositional reasoning and complex problem-solving abilities, reflecting higher-order cognition.
(3) OCR: TextVQA~\cite{singh2019towards} assesses text-rich visual question answering, evaluating the ability to recognize and reason over textual content in images.
(4) Hallucination: POPE~\cite{li2023evaluating} evaluates robustness against visual hallucinations and their consistency in visual grounding.
(5) Domain-specific: VisWiz~\cite{gurari2018vizwiz} targets real-world images taken by visually impaired users, evaluating the robustness of visual perception in domain-shifted scenarios.

\paragraph{Compared Methods. }
We compare Switch-KD against recent lightweight VLMs, including TinyLLaVA~\cite{zhou2024tinyllava}, Mini-Gemini~\cite{li2024mini}, SPHINX-Tiny~\cite{liu2024sphinx} and MobileVLM \cite{chu2024mobilevlm}, 
as well as distillation-based counterparts such as 
MoVE-KD~\cite{cao2025move}, LLaVA-MOD~\cite{shu2024llava}, LLaVA-KD~\cite{cai2024llava} and Align-KD \cite{feng2025align}.
For completeness, we also compare against SOTA VLMs, including LLaVA-1.5~\cite{liu2024improved} and Qwen-VL~\cite{bai2023qwen}.

\subsection{Benchmarked Results with the SoTAs}

As summarized in Table~\ref{tab:Switch-kd}, Switch-KD consistently achieves leading or competitive performance across models of different scales. 

For the large-scale group ($\ge$3B parameters), both TinyLLaVA-7B and TinyLLaVA-3B deliver strong overall results. Switch-KD-1.5B attains comparable $\text{Avg}_7$ scores (64.5 vs. 64.9 for TinyLLaVA-3B) with far fewer parameters, and notably surpasses the distilled MoVE-KD-v1.1-7B on key benchmarks such as MMBench (+4.0), VQAv2 (+1.5), and TextVQA (+0.7), underscoring its strong generalization capability.

In the mid-scale group ($\sim$1.5–2B parameters), Switch-KD-1.5B achieves leading performance over both distilled and non-distilled methods. Using similar LLM backbones, it outperforms recent distilled VLMs LLaVA-MoD and LLaVA-KD by 4.4 and 0.4 points in $\text{Avg}_7$, respectively, while maintaining greater data efficiency. Compared with the TinyLLaVA-1.5B baseline trained with a conventional PT-SFT scheme, Switch-KD achieves gains of 1.6 and 1.3 points in $\text{Avg}_7$ and $\text{Avg}_{10}$, respectively, demonstrating consistent improvements across benchmarks.

In the small-scale group ($<$1.5B parameters), Switch-KD-0.5B records a 4.0 points improvement in $\text{Avg}_7$ over SPHINX-Tiny while using only 8\% of its training data volume. It further outperforms LLaVA-MoD and LLaVA-KD of comparable scale by 3.7 and 1.1 points in $\text{Avg}_7$, respectively. Consistent with the 1.5B results, Switch-KD-0.5B also surpasses the TinyLLaVA baseline by 4.9 and 3.6 points in $\text{Avg}_7$ and $\text{Avg}_{10}$, respectively, highlighting its strength at lower model capacities.

Table~\ref{tab:align_kd_results} demonstrates Switch-KD's cross-architecture transferability. While Align-KD improves the baseline by 2.0 and 1.4 points with short and long instruction datasets, Switch-KD achieves a 3.5-point gain despite using a lighter backbone (Qwen2.5-1.5B vs. MobileLLaMA-1.7B) and one-third the data (1.2M samples), highlighting its distillation efficiency and robustness.

\subsection{Ablation Study and Analysis}

\begin{table*}[t]
\centering
\setlength{\tabcolsep}{4pt}
\renewcommand{\arraystretch}{1.15}
\begin{tabular}{lccccccccccc}
\toprule
\multirow{2}{*}{\textbf{Training Strategy}} & 
\multicolumn{3}{c}{\textbf{Percep. \& Underst.}} &
\multicolumn{4}{c}{\textbf{Cognition \& Reasoning}} &
\multicolumn{1}{c}{\textbf{OCR}} &
\multicolumn{1}{c}{\textbf{Specific}} &
\multicolumn{1}{c}{\textbf{Halluc.}} &
\multirow{2}{*}{\textbf{Avg\textsubscript{10}}}\\
\cmidrule(lr){2-4}\cmidrule(lr){5-8}\cmidrule(lr){9-9}\cmidrule(lr){10-10}\cmidrule(lr){11-11}
& MME & MMB & MMB$^{\text{CN}}$ & VQAv2 & GQA & SciQA & MMMU & TextVQA & VizWiz & POPE & \\
\midrule
TinyLLaVA-0.5B & 61.5 & 58.9 & 54.2 & 74.8 & 58.3 & \textbf{59.1} & \textbf{33.6} & 49.2 & 28.9 & 86.1 & 56.5 \\
w/o Switch & 63.1 & 60.8 & 57.3 & 77.8 & 59.7 & 58.1 & 31.7 & 51.0 & 41.5 & 87.0 & 58.8 \\
\rowcolor{green!10}
w/ Switch & \textbf{66.8} & \textbf{63.5} & \textbf{57.8} & \textbf{79.6} & \textbf{61.6} & 57.9 & 29.8 & \textbf{52.3} & \textbf{44.9} & \textbf{87.3} & \textbf{60.1} \\
\bottomrule
\end{tabular}
\caption{
Ablation study on the visual-switch architecture.
``w/o Switch'' refers to knowledge distillation with the standard alignment pathway, 
whereas ``w/ Switch'' incorporates the proposed visual-switch distillation.
}
\label{tab:visualswitch}
\end{table*}



\subsubsection{Is Visual‑Switch Distillation Effective?}
\label{ablation:switch}


We assess the effectiveness of the visual-switch architecture through both quantitative and qualitative analyses. 
As shown in Table~\ref{tab:visualswitch}, we compare the distilled performance of the 0.5B student model with and without the switch mechanism. 
Visual-switch distillation consistently outperforms both the baseline and the standard distillation architecture, raising the average score from 58.8 to 60.1. 
Notable gains are observed on VQAv2 (+1.8), GQA (+1.9), TextVQA (+1.3), and especially VizWiz (+3.4), suggesting that the switch effectively transfers the teacher’s robustness to challenging visual conditions such as low light and motion blur. 
Furthermore, improvements in the Perception \& Understanding ability (MME, MMB, and MMB$^{\text{CN}}$) and Hallucination robustness (POPE) indicate that the visual-switch mechanism primarily enhances cross-modal alignment and visual–semantic representation during distillation.

\begin{figure}[htp]
\centering
\begin{subfigure}[b]{0.22\textwidth}
    \centering
    \includegraphics[width=\linewidth]{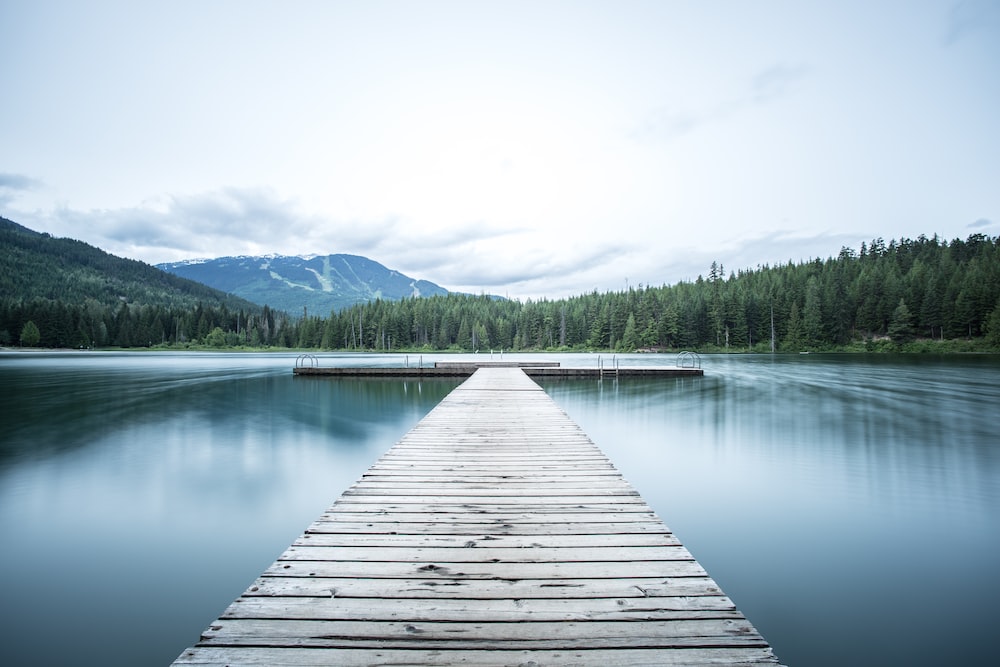}
    \caption{Original}
\end{subfigure}
\hfill
\begin{subfigure}[b]{0.22\textwidth}
    \centering
    \includegraphics[width=\linewidth]{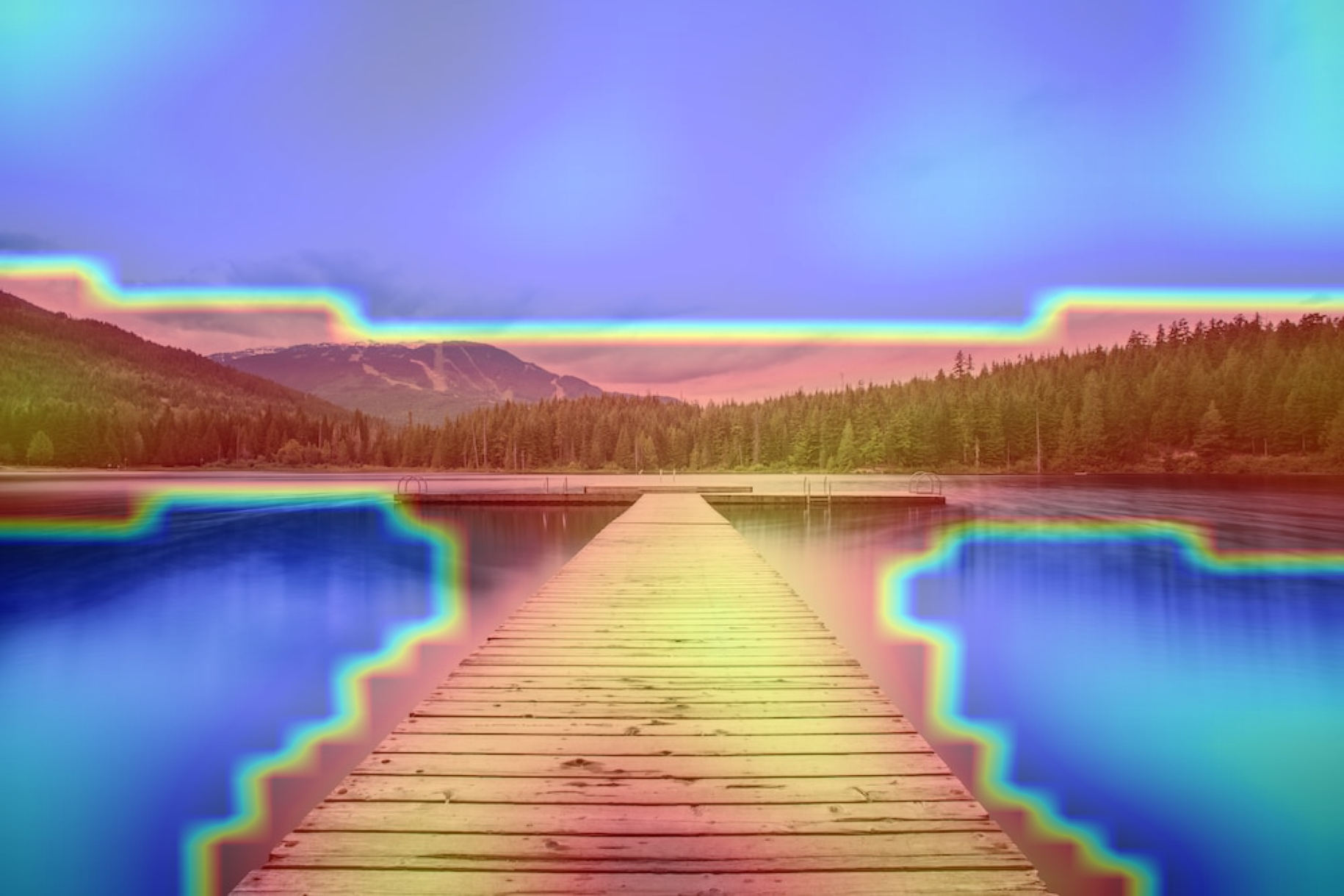}
    \caption{Teacher}
\end{subfigure}

\vspace{0.2em} 
\begin{subfigure}[b]{0.22\textwidth}
    \centering
    \includegraphics[width=\linewidth]{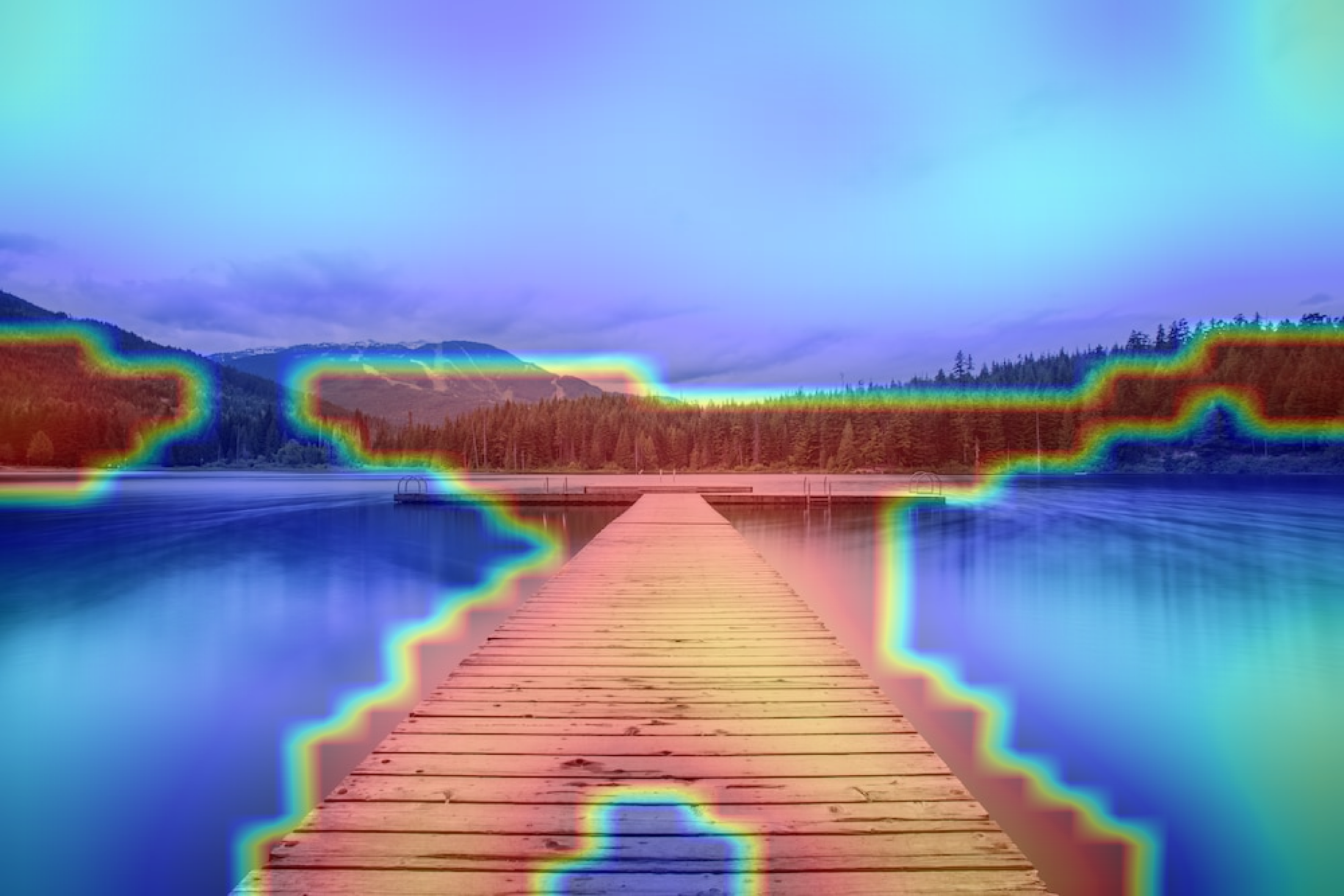}
    \caption{SFT}
\end{subfigure}
\hfill
\begin{subfigure}[b]{0.22\textwidth}
    \centering
    \includegraphics[width=\linewidth]{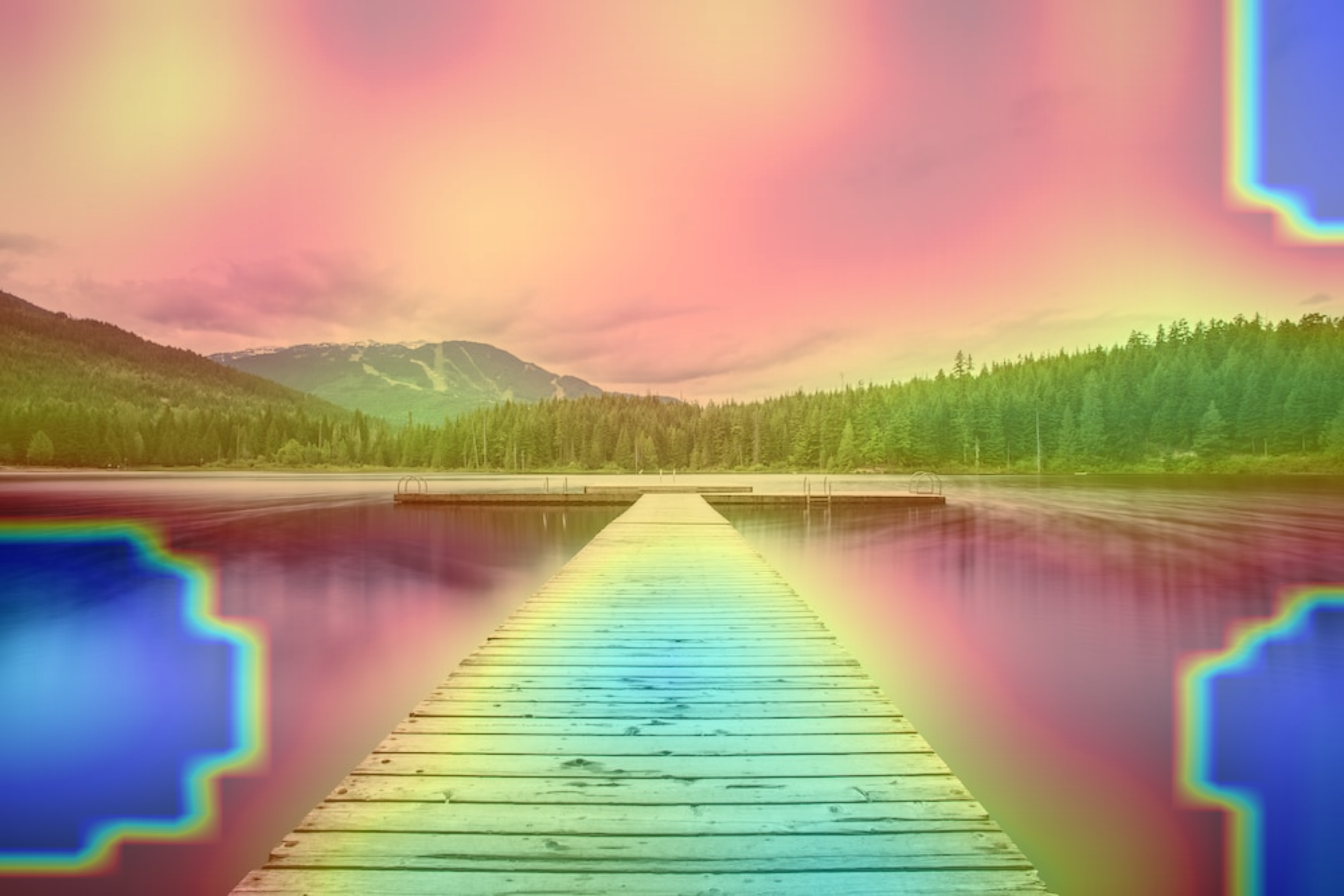}
    \caption{LLaVA-KD$^*$}
\end{subfigure}
\vspace{0.2em} 


\begin{subfigure}[b]{0.22\textwidth}
    \centering
    \includegraphics[width=\linewidth]{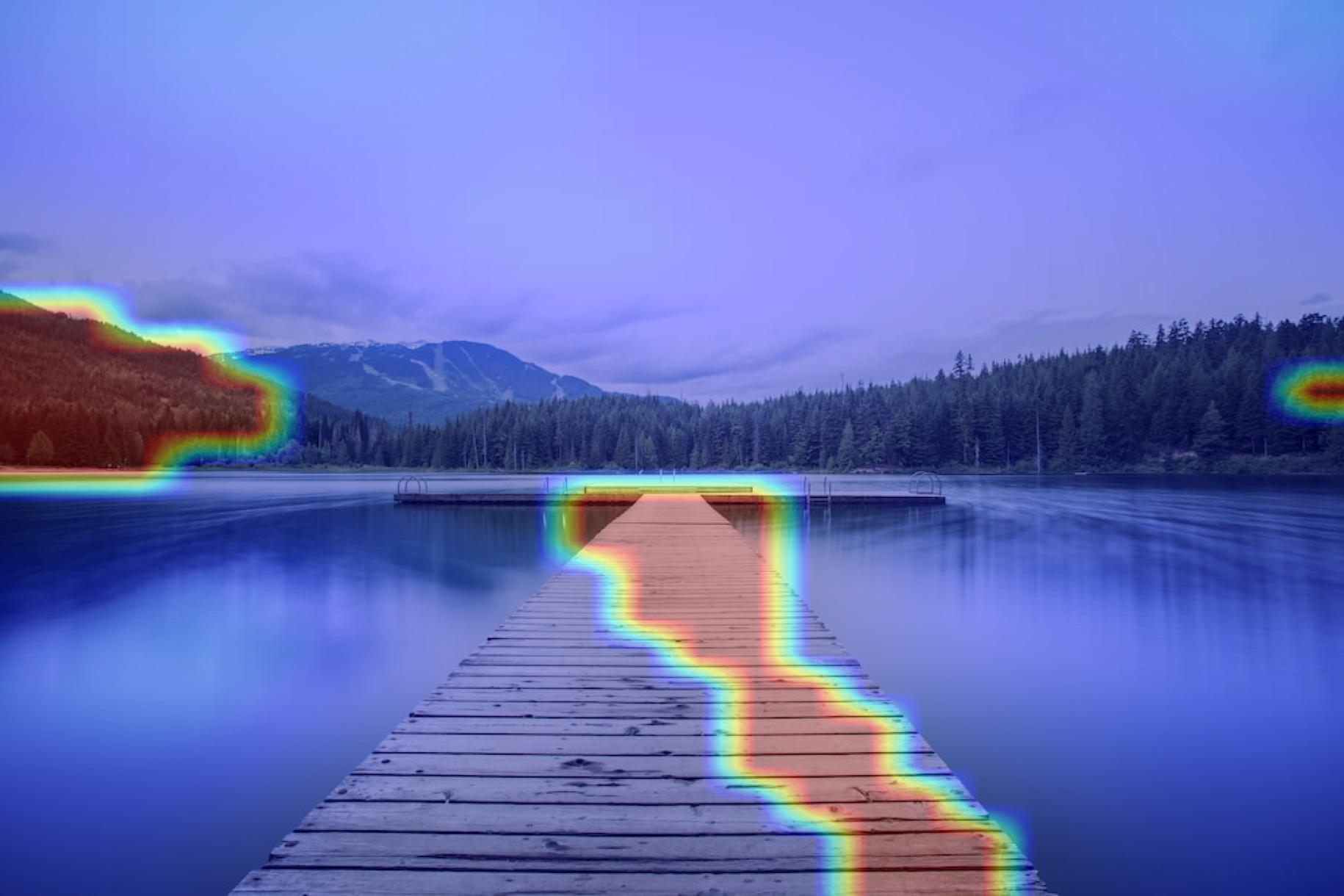}
    \caption{Align-KD$^*$}
\end{subfigure}
\hfill
\begin{subfigure}[b]{0.22\textwidth}
    \centering
    \includegraphics[width=\linewidth]{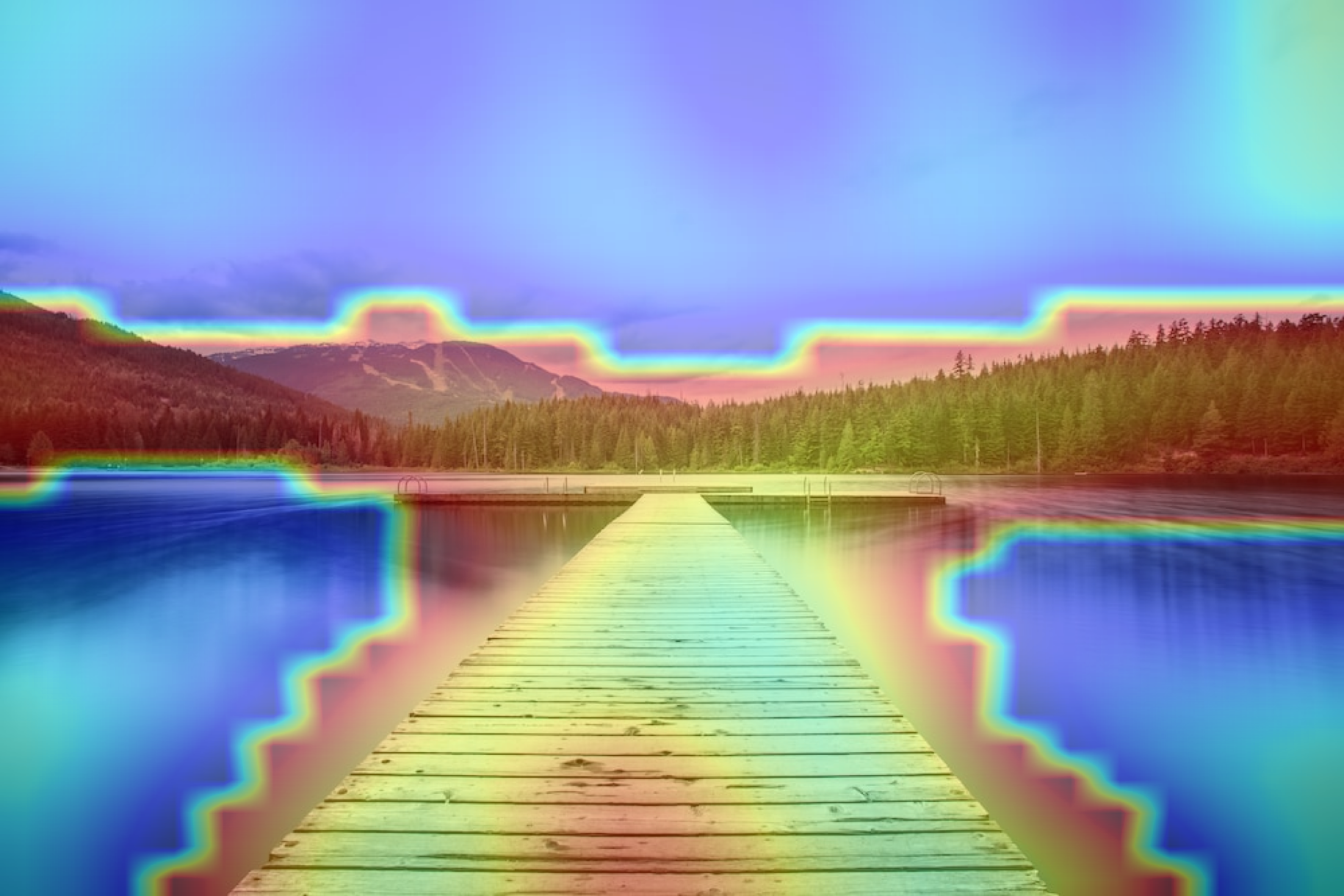}
    \caption{\textbf{Switch-KD}}
\end{subfigure}

\caption{
Visualization of attention maps.
\textbf{Switch-KD} (f) aligns the student’s visual focus with the teacher (b), producing attention maps consistent with teacher semantics. $^*$: visual encoder parameters frozen during training phases.
}
\label{fig:attention_map}
\end{figure}

Figure~\ref{fig:attention_map} compares attention maps from the teacher, an SFT baseline, two distillation methods, and our Switch-KD.
The teacher focuses on semantically critical regions (e.g., the intersection between a wooden bridge and distant mountains), demonstrating strong visual–semantic understanding.
While the SFT baseline approximates the teacher’s overall attention distribution, it fails to match the fine-grained semantic selectivity in specific regions.
In comparison, LLaVA-KD spreads attention almost uniformly across the image, lacking semantic focus, whereas Align-KD activates only partial semantic regions, further highlighting the limitations of modality-separated supervision under the frozen visual encoder strategy. 
By contrast, Switch-KD closely replicates the teacher’s attention pattern, attending to similar regions and achieving stronger visual–semantic alignment with the teacher.

\subsubsection{What Is the Impact of DBiLD Loss?}
\label{ablation:dbild-loss}

As shown in Table~\ref{tab:distill_strategy}, we fix all other variables and adopt a consistent standard distillation pipeline to examine the impact of different loss functions on performance. 
Using only the forward KL (FKL)~\cite{kim2016sequence} or reverse KL (RKL)~\cite{gu2023minillm} divergence yields the same average score of 58.3. 
Introducing the bidirectional difference loss into the FKL framework forms BiLD-FKL, which slightly raises the score to 58.4, confirming the effectiveness of bidirectional supervision. 
Building on this, replacing FKL with RKL in BiLD yields BiLD-RKL, providing a further 0.2 points gain and suggesting that emphasizing high-probability regions during optimization is beneficial. 
Finally, our proposed DBiLD‑RKL integrates dynamic Top‑$k$ selection, reaching an average score of 58.8 and outperforming the FKL baseline by 0.5 and BiLD by 0.4, demonstrating the efficiency and robustness of our dynamic bidirectional distillation loss design.

\begin{table*}[ht]
\centering
\setlength{\tabcolsep}{4pt}
\renewcommand{\arraystretch}{1.15}
\begin{tabular}{lccccccccccc}
\toprule
\multirow{2}{*}{\textbf{Distill Strategy}} & 
\multicolumn{3}{c}{\textbf{Percep. \& Underst.}} &
\multicolumn{4}{c}{\textbf{Cognition \& Reasoning}} &
\multicolumn{1}{c}{\textbf{OCR}} &
\multicolumn{1}{c}{\textbf{Specific}} &
\multicolumn{1}{c}{\textbf{Halluc.}} &
\multirow{2}{*}{\textbf{Avg\textsubscript{10}}}\\
\cmidrule(lr){2-4}\cmidrule(lr){5-8}\cmidrule(lr){9-9}\cmidrule(lr){10-10}\cmidrule(lr){11-11}
& MME & MMB & MMB$^{\text{CN}}$ & VQAv2 & GQA & SciQA & MMMU & TextVQA & VizWiz & POPE &\\
\midrule
FKL & 63.2 & 60.4 & 55.8 & 77.0 & 59.8 & 58.1 & 32.3 & 49.5 & 39.9 & 86.6 & 58.3 \\
RKL & 63.3 & 60.3 & 57.0 & 76.7 & 59.4 & 58.5 & 32.4 & 49.8 & 39.0 & 86.7 & 58.3 \\
BiLD-FKL & 63.4 & 60.7 & 56.1 & 77.3 & 59.5 & 58.1 & 31.3 & 50.3 & 40.3 & 86.7 & 58.4 \\
BiLD-RKL & \textbf{63.5} & \textbf{60.9} & 56.8 & 77.0 & 59.3 & 58.0 & 31.0 & \textbf{51.2} & 41.5 & 86.3 & 58.6 \\
DBiLD-FKL & 63.3 & 60.7 & 56.6 & 77.4 & 59.5 & \textbf{58.4} & 31.3 & \textbf{51.2} & 40.9 & \textbf{86.9} & 58.6 \\
\rowcolor{green!10}
DBiLD-RKL & 63.1 & 60.8 & \textbf{57.3} & \textbf{77.8} & \textbf{59.7} & 58.1 & \textbf{31.7} & 51.0 & \textbf{41.5} & 87.0 & \textbf{58.8} \\
\bottomrule
\end{tabular}
\caption{
Ablation study of different distillation strategies.
BiLD-FKL denotes the original BiLD loss using forward KL divergence, while BiLD-RKL replaces FKL with reverse KL divergence. 
DBiLD-FKL applies dynamic Top-$k$ selection to BiLD-FKL, and DBiLD-RKL combines reverse KL divergence with dynamic Top-K selection as our final choice.
}
\label{tab:distill_strategy}
\end{table*}

\begin{table}[ht]
\centering
\setlength{\tabcolsep}{12pt}
\renewcommand{\arraystretch}{1.15}
\begin{tabular}{lcc}
\toprule
\textbf{Training Scheme} & \textbf{$\text{Avg}_{7}$} & \textbf{$\text{Avg}_{10}$} \\
\midrule
PT-SFT   & 52.9&56.5 \\
DPT-SFT  & 54.1&57.4 \\
PT-DFT   &55.9& \textbf{58.8} \\
DPT-DFT  &\textbf{56.0} & 58.7 \\
\bottomrule
\end{tabular}
\caption{
Ablation study on different training schemes.
}
\label{tab:training_scheme}
\end{table}

\begin{table}[t]
\centering
\setlength{\tabcolsep}{6pt}
\renewcommand{\arraystretch}{1.15}
\begin{tabular}{l l c c}
\toprule
\textbf{Teacher LLM}& \textbf{Student LLM}&  \textbf{$\text{Avg}_{7}$} &  \textbf{$\text{Avg}_{10}$} \\
\midrule
\multirow{2}{*}{Qwen2.5-3B} & Qwen2.5-0.5B & 57.8 & 60.1 \\
& Qwen2.5-1.5B & 63.8 & 64.8 \\
\midrule
\multirow{2}{*}{Qwen2.5-7B} & Qwen2.5-0.5B & 57.1 & 59.8 \\
& Qwen2.5-1.5B & 64.5 & 65.3 \\
\bottomrule
\end{tabular}
\caption{
Ablation study on teacher models with different sizes.
}
\label{tab:teacher_student}
\end{table}

\subsubsection{Why Distill Only in Fine‑Tuning?}
\label{ablation:dft}
We further evaluate the impact of applying distillation at different stages of the training pipeline.
As shown in Table~\ref{tab:training_scheme}, the baseline PT–SFT follows the standard two-stage process of Pre-Training and Supervised Fine-Tuning. 
Introducing distillation during Pre-Training (DPT–SFT) yields a modest 0.9 points gain on $\text{Avg}_{10}$, suggesting that early-stage supervision provides limited improvement. 
In contrast, distilling only during Fine-Tuning (PT–DFT) brings a larger 2.3 points gain, indicating that task-specific imitation of the teacher’s outputs facilitates more effective knowledge transfer. 
However, applying distillation at both stages (DPT–DFT) results in a slight 0.1 points drop, showing no additive benefit and confirming that the main improvement comes from replacing conventional SFT with DFT. 
To prioritize training efficiency, we avoid LLaVA-KD’s three-stage framework (DPT–SFT–DFT), adopting instead a streamlined two-stage pipeline that maximizes distillation performance while avoiding additional resource costs.

\subsubsection{How Does Scaling Affect Distillation?}
\label{ablation:scale}
In Table~\ref{tab:teacher_student}, we additionally analyze the impact of teacher model size on distillation performance under different teacher–student configurations. 
When using Qwen2.5-3B as the teacher LLM, the student models achieve average scores of 60.1 and 64.8 for the 0.5B and 1.5B variants, respectively. 
Replacing the teacher LLM with a larger Qwen2.5-7B yields further gains, where the 1.5B student reaches 65.3 on $\text{Avg}_{10}$, surpassing its 3B-teacher counterpart by 0.5 points. 
However, for the 0.5B student, using a stronger teacher causes a slight performance drop (59.8 vs. 60.1), indicating that the benefits of larger teachers are constrained and can even be reversed by limited student capacity. 
This observation reinforces the overall finding that, while larger teachers provide better supervision, the effectiveness of distillation is ultimately bounded by the representational capacity of the student model.


\subsubsection{Where Can Switch‑KD Be Improved?}
\label{ablation:further}
Note that Switch-KD currently requires feature-space and vocabulary consistency between teacher and student for stable cross-modal knowledge transfer. While effective, this limits its applicability to heterogeneous architectures. Future work may investigate architecture-agnostic distillation or adapter-based mappings to relax this constraint, enabling knowledge transfer across diverse VLM designs.
\section{Conclusion}
\label{sec:con}


In this work, we introduce Switch-KD, a unified vision–language knowledge distillation framework with two key components: a visual-switch architecture and a Dynamic Bi-directional Logits Difference (DBiLD) loss. 
The visual-switch architecture switches the student’s visual outputs into the teacher’s language pathway, enabling implicit cross-modal supervision within a shared text-probability space. 
The DBiLD loss adaptively selects the most informative top-$k$ logits and performs bi-directional rank alignment via reverse KL divergence, focusing on high-confidence regions while preserving the teacher’s relative probability structure. 
Experiments on ten multimodal benchmarks show that Switch-KD consistently surpasses state-of-the-art distillation and lightweight VLM approaches. 
We believe it offers a practical solution for deploying high-performance vision–language models in resource-constrained scenarios and may inspire future research on unified multimodal distillation.
{
    \small
    \bibliographystyle{ieeenat_fullname}
    \bibliography{main}
}

\end{document}